\documentclass{article}
\usepackage{spconf,amsmath,graphicx, animate}
\usepackage{tabularx}
\usepackage{hyperref}
\usepackage{url}
\usepackage{pdfpages}
\usepackage{graphicx, animate}
\usepackage{times}
\usepackage{epsfig}
\usepackage{amsmath}
\usepackage{algorithm}
\usepackage{algpseudocode}
\usepackage{setspace}
\usepackage[cal=euler]{mathalfa}
\usepackage{float}
\usepackage[caption = false]{subfig}

\usepackage{xcolor}
\usepackage{pifont}
\hypersetup{
    colorlinks=true,
    linkcolor=teal,
    citecolor=violet,
    filecolor=magenta,      
    urlcolor=violet,
}
\usepackage{pifont}

\definecolor{newcolor}{rgb}{.8,.349,.1}
\DeclareMathOperator*{\argmax}{arg\,max}
\hypersetup{colorlinks=true,citecolor=green}

\title{Learning Class Regularized Features for Action Recognition}
\name{\normalsize Alexandros Stergiou$^{ \dagger}$\thanks{$^{\dagger}$Corresponding author}, Ronald Poppe, Remco C. Veltkamp}

\address{
Department of Information and Computing Sciences, Utrecht University, Utrecht, Netherlands\\
\normalsize \{a.g.stergiou,  r.c.veltkamp,  r.w.poppe\}@uu.nl}
\begin{document}
%\ninept
%
\maketitle

\begin{abstract}
Training Deep Convolutional Neural Networks (CNNs) is based on the notion of using multiple kernels and non-linearities in their subsequent activations to extract useful features. The kernels are used as general feature extractors without specific correspondence to the target class. As a result, the extracted features do not correspond to specific classes. Subtle differences between similar classes are modeled in the same way as large differences between dissimilar classes. To overcome the class-agnostic use of kernels in CNNs, we introduce a novel method named \textit{Class Regularization} that performs class-based regularization of layer activations. We demonstrate that this not only improves feature search during training, but also allows an explicit assignment of features per class during each stage of the feature extraction process. We show that using \textit{Class Regularization} blocks in state-of-the-art CNN architectures for action recognition leads to systematic improvement gains of 1.8\%, 1.2\% and 1.4\% on the \textit{Kinetics}, \textit{UCF-101} and \textit{HMDB-51} datasets, respectively.
\end{abstract}

\begin{keywords}
Regularization, explainable convolutions, spatio-temporal activations. 
\end{keywords}
%

%\linenumbers
%%%%%%%%% Introduction
\section{Introduction}
\label{Section1}

\begin{figure}[ht]
\centering
\includegraphics[width=0.85\linewidth]{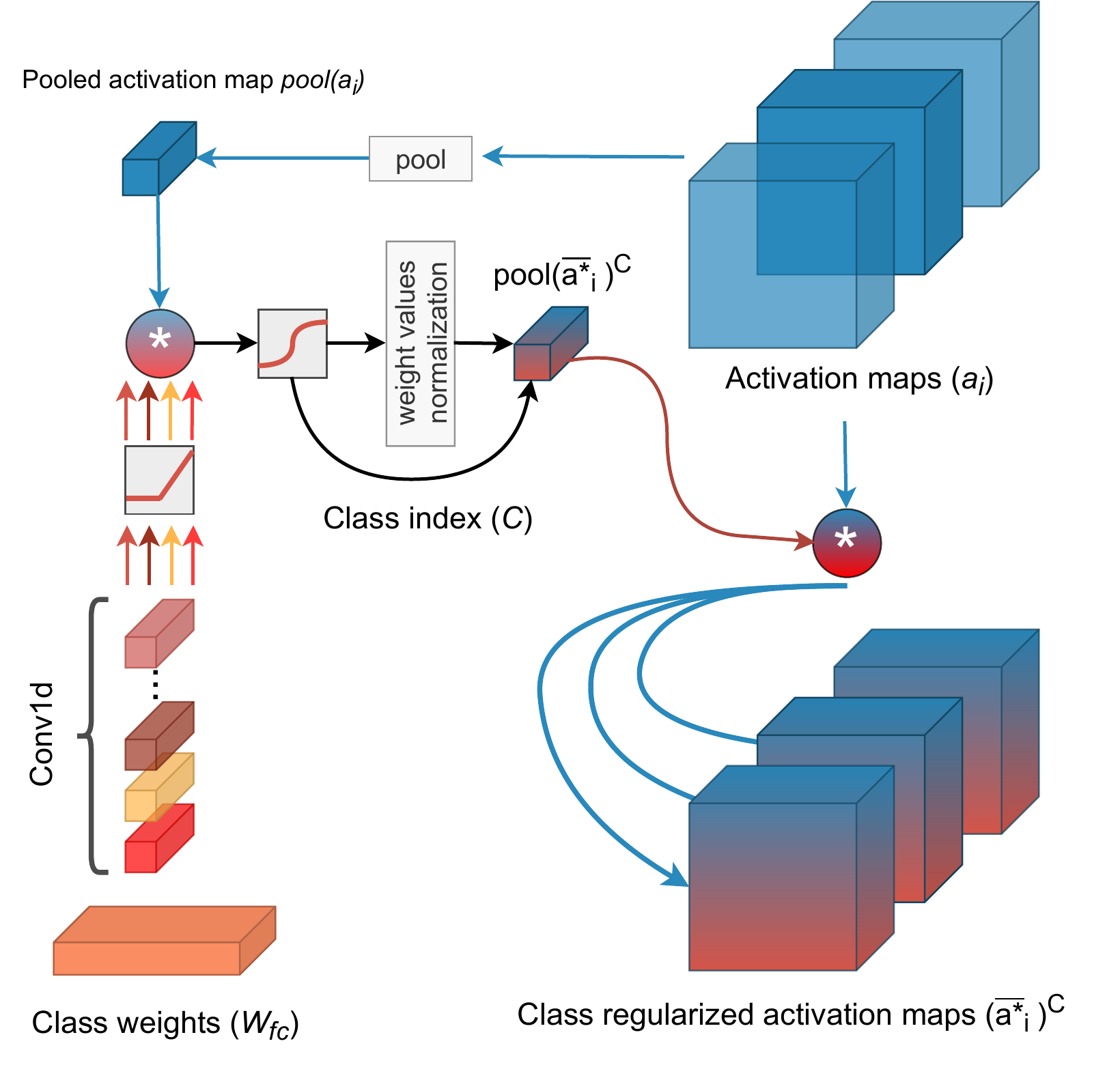}%
\caption{\textbf{Class Regularization}. Activation maps are vectorized (pool($a_i$)) and multiplied by the convolved class weights (for dimensionality matching), to select the resulting highest class activation (pool$(\overline{a^{*}_{i}})^{C})$), and regularize the layer activations.}
\label{fig:classreg}
\end{figure}

% Action recognition in the context of Deep Learning.
Video-based action recognition has seen tremendous progress since the introduction of Convolutional Neural Networks (CNNs). The use of multiple 3D convolutional operations in each layer has shown to effectively capture informative and descriptive spatio-temporal features. A large body of work has focused on finding optimal architectures, depth and feature-extraction methods \cite{herath2017going,stergiou2019analyzing}.

% The use of multiple 3D convolutional operations in each layer has been shown to effectively capture informative and descriptive spatio-temporal features

%\begin{figure*}[ht]
%    \subfloat[Network view]{{\includegraphics[width=0.6\linewidth,valign=t]{Class_Reg_pipeline.pdf}}}%
%    \qquad
%    \subfloat[\vspace*{-12.5mm} Block view]{{\includegraphics[width=0.33\linewidth,valign=t]{Class_Reg_block.pdf}}}%
%   \caption{\textbf{Class Regularization}. An additional pathway between class weights and intermediate features is added (a) connecting different parts of the network across iteration steps. In \textit{Class Regularization} modules, activation maps are pooled to vector volumes (pool($a_i$)) and multiplied by the class weights in order to select the resulting highest class activation (pool$(\overline{a^{*}_{i}})^{C})$). An overview of the in-block operations (b) can be found in Algorithm~\ref{algo:algo1}.}
%\label{fig:classreg}
%\end{figure*}

% Feature inclusion on class predictions
In recognition tasks, networks include multiple layers that are stacked together in a single, hierarchical architecture. Features are extracted through successive convolution operations, where each layer employs a set of kernels whose parameters are learned during training. Early layer kernels focus on simple textures and patterns, while deeper layers focus on complex object parts or specific parts of scenes. However, as these features become more dependent on the different weighting of neural connections in previous layers, only a portion of them becomes descriptive for a specific class \cite{bau2019visualizing,gilpin2018explaining}. Yet, all kernels are learned in a class-agnostic way. This hinders easy interpretation of the part of the network that is informative for a specific class. Moreover, it complicates model transfer to other datasets.

We explicitly focus on this space-time relationship and propose a method named \textit{Class Regularization}. We relate class information to extracted features of different network blocks. This information is added back to the network as a means of amplifying activation values, with respect to predicted classes. \textit{Class Regularization} has a beneficial effect on the non-linearities of the network by decreasing or increasing the effects of the activations. Based on this, the architecture can effectively distinguish between the most class-informative kernels in each part of the network hierarchy given a selected class. This also reduces the dependency on many uncorrelated features during the final class predictions, essentially penalizing overfitting given the random sampling noise of the data.\pagebreak

Our contributions are the following:
\begin{itemize}
  \item We propose \textit{Class Regularization}, a regularization method applied in spatio-temporal CNNs without changing the overall network structure.
  
  \item We introduce a weight sharing function for learned weights of previous epochs with \textit{Class Regularization}.
  
  \item We demonstrate the improvement in model explainability through intermediate class-spceific features.
  
% without the need of back-propagating from the predictions to a particular layer in order \citep{stergiou2019class},
  
  \item We report performance gains for benchmark action recognition datasets \textit{Kinetics}, \textit{UCF-101} and \textit{HMDB-51} by including \textit{Class Regularization} blocks.
  
 % We further perform a McNemar's test \citep{mcnemar1947note} to directly compare architectures with and without \textit{Class Regularization}.
\end{itemize}

% Overview
The advances made in vision-based action recognition are discussed in Section~\ref{sec:sec2}. A detailed overview of the algorithm appears in Section~\ref{sec:sec3}. Results and evaluation tests are presented in Section~\ref{sec:sec4}. We conclude in Section~\ref{sec:sec5}.

\section{Related work}
\label{sec:sec2}

% Hand-crafted feature representation
%Actions in videos can be represented as volumes spanning across a time dimension. Initially, research efforts in the field considered feature descriptors such as Histogram of Oriented Gradients (HOG) which were also extended to include temporal dimensions \cite{klaser2008spatio}, Histogram of Optical Flow and Motion Bounded Histograms (MBH). Later works \cite{wang2011action,wang2013dense} have also combined multiple feature descriptors for discovering individual point trajectories over videos, named Dense Trajectories (DT). Deformable Parts Models (DPMs) \cite{felzenszwalb2010object} have further empowered the recognition of actions in videos, by modeling people descriptors as sets of parts of with multiple deformations between them, which also allowed distinguishing between similar classes.

% Using Deep Learning Architectures for videos

%Although hand-coded feature descriptors were able to obtain notable results in many popular datasets, they were limited towards their high correspondence to the features selected. 

%With the introduction of deep neural approaches that were based on the discovery of informative features in a hierarchical fashion \cite{he2016deep, simonyan2014very}, additionally significant advancements have been made in the recognition of actions in videos as these architectures provided the templates for adaptions to also accommodate temporal information.

% Appearance and motion as individual source of information.
Because of the indirect relationship between temporal and spatial information, one of the first attempts on video recognition with neural models was the use of Two-stream networks \cite{simonyan2014two}. These models contain two separate branches for still video frames and optical flow inputs, respectively. Two-steam networks were also used as a base method for approaches such as Temporal Segment Networks (TSN) \cite{wang2016temporal} using scattered snippets from the video and later fusing their predictions. This also led to research on the selection of frames \cite{diba2017deep} while other approaches use residual connections \cite{feichtenhofer2016spatiotemporal} to share spatio-temporal information across multiple layers.

% Incorporating motion in convolutions.
Other approaches consider 3D convolusions, which include time information as part of their operations and have shown to outperform standard image-based networks in video classification. A fusion of Two-stream networks and 3D convolutions has been explored with the I3D architecture \cite{carreira2017quo}, with two spatio-temporal models trained in parallel on both frame and optical flow data. Further structures include Residual Networks \cite{hara2018can}, depth-wise and channel-wise convolutions to deal with spatio-temporal data \cite{chen2018multifiber, tran2019video}, combinations of spatial-only followed by temporal-only filters \cite{qiu2017learning,tran2018closer} and the use of long-sequence and short-sequence kernels \cite{feichtenhofer2019slowfast}. 

% Lack of class-specific information
Although these techniques have shown great promise, there is still a lack of better spatio-temporal representations for intermediate network layers. Yet, no standardized way for processing the temporal information exists. Our proposed method, named \textit{Class Regularization}, can be added to networks with minimum additional computational costs in order to further enforce the relation between features and action classes.

% Regularization in NNs

\section{Regularization for convolutional blocks}
\label{sec:sec3}
Explicitly adding class information through regularization is challenging based on the ambiguity of the model's inner workings. The underlying idea is that in each layer, different combinations of extracted spatio-temporal features lead to patterns that are significant parts of different classes. These patterns are depth dependent, i.e. deeper layers can distinguish class-specific features better given their higher feature complexity. Therefore, class estimates at different parts of the model should be weighted differently. We define an \textit{affection rate} value ($A$), that specifies how strong the intermediary predictions in that layer should be. The values are chosen given the layer depth and the level of uncertainty of their class estimates. We further use point-wise convolutions for feature dimensionality matching between the predictions and layer activations.\\
We now discuss the various steps that layer activations are regularised over predicted class weights as shown in Figure~\ref{fig:classreg}

%\begin{spacing}{0.8}
%\begin{algorithm}
%\setstretch{1.3}
%\caption{Class regularization overview}
%\label{algo:algo1}
%\begin{algorithmic}[1]
%
%\algrenewcommand\algorithmicprocedure{\textbf{Input: }}
%
%\Inputs{Values of activation map $a_{i}$ for $i^{th}$ %layer.}
%\Outputs{Class-regularized activations %$(\overline{a^{*}_{i}})^{C}$}
%\item[]
%%\State $pool(a_{i})\gets \sum \limits_{z=0}^{kZ-1} \sum %\limits_{y=0}^{kY-1} \sum \limits_{a=0}^{kX-1} %\frac{a_{i}(M_{i},s * f + z, s * h + y, s * w + %a)}{kZ+kY+kX} $
%%\par\hskip\algorithmicindent
%%\Comment{Sampling operation for $a_{i}$ activation map}
%\State $W_{i} \gets relu(W_{fc} * \mathcal{W})$
%\par\hskip\algorithmicindent \Comment{Weight dimensionality conversion}

%\State $Z_{i} \gets W_{i} * pool(a_{i})$
%\par\hskip\algorithmicindent \Comment{Weighted sum per class neuron}

%\State  $C \gets \underset{j}{\argmax} \{S(Z^{[1]}_{i}),...,S(Z^{[CL]}_{i})\} \: \forall \: S(Z^{[j]}_{i}) = \frac{\epsilon^{Z^{[j]}_{i}}}{\sum \limits_{c \in {1,...,CL}} \epsilon^{Z^{[c]}_{i}}}$
%\par\hskip\algorithmicindent
%\Comment{Largest softmax activation class search}

%\State $\widehat{W}_{i} \gets \mathfrak{A} * \frac{(W_{i} - min\{W_{i}) * (1-\mathfrak{A})}{max\{W_{i}\} - min\{W_{i}\}}$
%\par\hskip\algorithmicindent \Comment{Weight scaling}

%\State $(\overline{a^{*}_{i}})^{C} \gets %\widehat{W}^{[c]}_{i} * a_{i}$
%\par\hskip\algorithmicindent \Comment{Final weight %regularization.}
%\end{algorithmic}
%\end{algorithm}
%\vspace{-5mm}
%\end{spacing}

\begin{figure*}[ht]
\includegraphics[width=\textwidth]{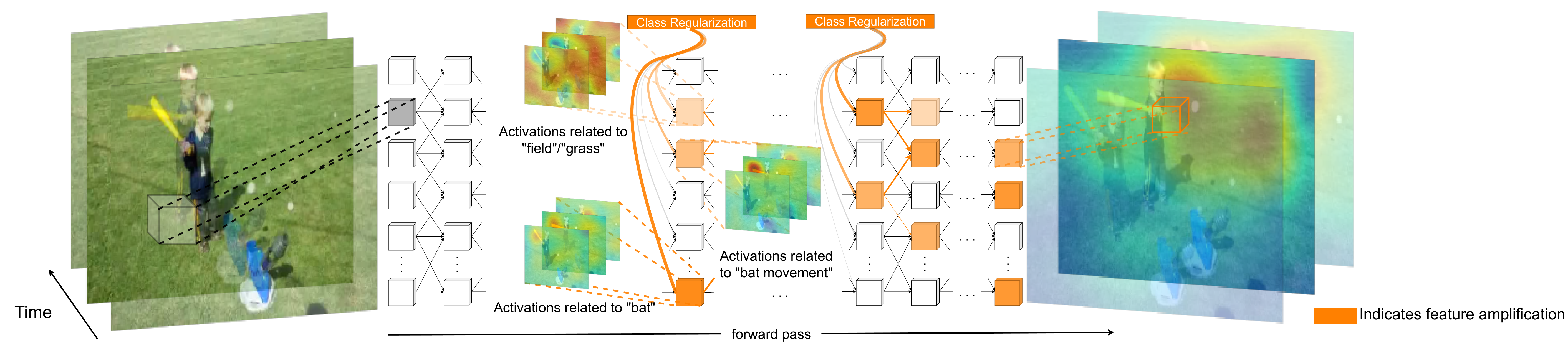}
\caption{Visualization of feature amplification. As class specific saliency is re-used by the network, informative spatio-temporal features for specific classes during an iteration will be amplified. The effect of this amplification is propagated to deeper layers in the network through the connections of the layers in which \textit{Class Regularization} is applied.}
\label{fig:layer_feature_wise_saliency}
\vspace{-5mm}
\end{figure*}

\subsection{Layer fusion with class predictions}
\label{sec:section3.1}

% Sampling activation maps.
Class estimates through features from convolution block ($i$) are obtained by initially creating a vector representation of the activations channels. Considering the produced activation map of the $i^{th}$ block (denoted as $a_{i}$) and spatio-temporal sampling operation, $pool(a_{i})$ (Equation~\ref{eq:equation1}), the produced volume can be interpreted as a descriptor containing feature intensity values.

\begin{equation}
\label{eq:equation1}
pool(a_{i}) = \frac{1}{F \times H \times W}\sum \limits_{f=0}^{F} \sum \limits_{h=0}^{H} \sum \limits_{w=0}^{W} a_{(f,h,w,i)}
\end{equation}

% Class weight insertion 
Class predictions are obtained based on the class weights of the network's classifier ($W_{fc}$), as updated by the preceding iteration. Thus allowing to establish a relationship between the previous and current iterations, in a recurrent fashion. As the feature space of the $i^{th}$ layer's activations varies from the prediction weights, a 3D point-wise convolution ($W_{i}=conv(W_{fc})$) is applied to the class weights.

% Class predictions based on example feature vector descriptor.
Based on the vectorized activations and class weights, the produced class activation volume ($Z_{i}=W_{i} * pool(a_{i})$) will be of size $\{Z^{[1]}_{i},...,Z^{[CL]}_{i}\}$ with ($CL$) being the number of classes. This operation allows an early estimate for the indexes of the most relevant features for each class.

% Flaws of using predictions from layers in the loss function. 
%It is possible to use the outputs of the multiplication operation as part of a separate loss function. We have two main reasons to refraining from doing so. First, based on the limited feature complexity in early layers of the network, educated class predictions are significantly more difficult to be made. This also corresponds to the notion of hierarchical feature extraction in CNNs. Second, through these multiple output points, multiple error derivatives are to be calculated which will slow down the training process significantly. This would corresponds to $L + 1 - l$ updates for layer $l$ of a $L$ block network.

\subsection{Class-specific excitation through class estimates}
\label{sec:section3.2}

% Information sub-sampling and finding max class probability.
%To incorporate class-specific information extracted by weighting every feature ($d$) of the pooled activation map ($pool(a_{i})$), we use the weight vector ($W^{[c]}_{i}$) that corresponds to the class ($[c]$) of the highest Softmax unit ($s^{[c]}_{i}$) of the group as in Algorithm~\ref{algo:algo1}.

% Main mindset
Considering the class-based activations ($Z_{i}$), the maximum class probability ($C$) can then be obtained through a normalized exponential function as in Equation~\ref{eq:equation2}. This converts the weighed sum logit score to a probabilistic distribution over all classes ($S(Z_{i})$).The obtained maximum class probability index ($C$) can be then used to select a specific class weight ($W^{[C]}_{i}$) based on which the activations of the layer will be regularized. 

\begin{equation}
\label{eq:equation2}
\resizebox{.85\hsize}{!}{$C \gets \underset{j}{\argmax} \{S(Z^{[1]}_{i}),...,S(Z^{[CL]}_{i})\} \: \forall \: S(Z^{[j]}_{i}) = \frac{\epsilon^{Z^{[j]}_{i}}}{\sum \limits_{c \in {1,...,CL}} \epsilon^{Z^{[c]}_{i}}}$}%
\end{equation}

%weight vector that is based on the feature vector, can be seen as an efficient approximation for the most informative features/channels in the activation map volume. This allows for weighting each feature differently based on how informative it is, given input $a_{i}$. This does not directly effecting the convolutional operations and allows to method to be used as an add-on. 

% Using alpha 
For amplifying each of the features of the activations, the selected class weights selected ($W^{[C]}_{i}$) are normalized within a discrete range of values. This is done for scaling down the effects of features, that are less informative for a specific class, while informative features are scaled up. This is performed based on the \textit{affection rate} value ($A$) that determines the bounds that the weight vector will be normalized to ($\widehat{W}^{[C]}_{i}$): as in Equation~\ref{eq:equation3}.

\begin{equation}
\label{eq:equation3}
\widehat{W}^{[C]}_{i} \gets A * \frac{(W^{[C]}_{i} - min\{W^{[C]}_{i}) * (1-A)}{max\{W^{[C]}_{i}\} - min\{W^{[C]}_{i}\}}
\end{equation}

% Shortcomings of zero-mean normalization and reasons for making the mean 1.
We are not using a standardization method as in \textit{batch normalization} \cite{ioffe2015batch} that guarantees a zero-mean output. This is because we use a multiplication operation for including the class weight information to the activation maps. Therefore, zero-mean normalization will remove part of the information as values below one will decrease the feature intensity. It also hinders performance as it effectively contributes to the occurrence of \textit{vanishing gradients} with the produced activation map values being reduced at each iteration.

% Class information injection in activation maps
In our final step, we inflate the normalized weight vector ($\widehat{W}^{[c]}$), to correspond to the same dimensions as the spatio-temporal activation maps and, in turn, create class-excited activations ($\overline{a^{C}_{i}}=\widehat{W}^{[C]}_{i} * a_{i}$)

\subsection{Improving visual explainability}
\label{sec:section3.3}

% Visualising individual blocks predictions with Class Reg.
%As \textit{Class Regularization} is based on the injection of class-based information inside the feature-extraction process. A direct correlation between classes and features is made at each block in which the method is applied. 
Being able to represent the class features, given a different feature space, further empowers the overall explainability capabilities of the model. Through feature correlation, the method alleviates the curse of dimensionality problem of current visualization methods that rely on back-propagating from the predictions to a particular layer \cite{stergiou2019class}. Since the classes are represented in the same feature space as the activation maps of the block, we can discover regions in space and time that are informative over multiple network layers. To the best of our knowledge, this is the first method to visualize spatio-temporal class-specific features at each layer of the network. This can be seen in Figure~\ref{fig:layer_feature_wise_saliency} through the extension of the \textit{Saliency Tubes}\cite{stergiou2019saliency} method, for each block, to create visual representations of the features with the highest activations per class.

\section{Experiments}
\label{sec:sec4}
%We demonstrate the merits of \textit{Class Regularization} on the action recognition classification performance on benchmark datasets, and using a number of widely used CNN architectures. We further statistically compare the classification performance between predictions from different architectures and from different blocks within the same architecture through a McNemar's statistical significance test.

% Datasets and TL stetup
We demonstrate the merits of \textit{Class Regularization} on three widely used datasets: Kinetics-400 \cite{kay2017kinetics}, UCF-101 \cite{soomro2012ucf101} and HMDB-51 \cite{kuehne2011hmdb}. The models trained on Kinetics are initialized with a standard Kaiming initialization, without inflating the 3D weights. This was done to allow for a direct comparison between architectures with and without \textit{Class Regularization} blocks and compare the respective accuracy rates in each case. For all the experiments we use \textit{SGD} as our optimizer with 0.9 momentum. \textit{Class Regularization} is added at the end of each bottleneck block in the ResNet architectures and at the end of each mixed block in I3D. 

%We use transformations for both space and time dimensions. In the time dimension, the clips used are 16 frames long and randomly extracted from different sub-sequences of the video. For the validation sets, we only use the center $n$ frames (where $n$ is the number of selected frames). Spatially, we use a cropping size of $112\times112$ and $256\times256$ for fine-tuning.

% Datasets used
%\textbf{Datasets.} Kinetics-400 consists of roughly 240K training videos and 20k validation videos of 400 different human actions. All models are trained for 170 epochs. We report the top-1 accuracy alongside the computational cost (FLOPs) for each of the networks using spatio-temporally cropped clips. 
%UCF-101 and HMDB-51 have 13k and 7k videos, respectively. They are used to demonstrate the transfer abilities of the proposed \textit{Class Regularization} as well as the usability of our method in smaller datasets.

\subsection{Main results}

% Accuracy to previously reported models
A comparison between our results on Kinectics-400 and those previously reported in literature appears in Table~\ref{table:table1}. Existing networks consider a complete change in the overall architecture or convolution operations in models, which is significantly computationally challenging given the large memory (based on batch sizes) and computations requirements of spatio-temporal models (as shown by the number of GFLOPs). New models need to be trained for a significant number of iterations in order to achieve mild improvements: +3.6\% from I3D \cite{carreira2017quo} to R(2+1)D \cite{tran2018closer}, while additionally pre-training on even larger datasets \cite{ghadiyaram2019large}. In contrast, the proposed \textit{Class Regularization} method is used on top of existing architectures and only requires fine-tuning the dimensionality correspondence between the number of features in a specific layer and the features that are used for class predictions. For a direct comparison, in the retrained models with batch sizes of 32, we achieve an overall average improvement of: +1.29\% on 101-layer ResNet, +1.5\% on 50-layer Wide ResNet and +1.45\% on I3D as seen in Tables~\ref{table:table1},~\ref{table:table2}.

\begin{table}[htb]
\vspace{-2mm}
\caption{Comparisons of accuracy rates over different spatio-temporal convolutional architectures on Kinetics-400. Computational overhead is denoted by the number of GFLOPS. \label{table:table1}}
\vspace{-6mm}
\begin{center}
\resizebox{\columnwidth}{!}{
\begin{tabular}{l|c|c|c|c}
\hline
\textbf{Model} & \textbf{Pre-training} & \textbf{Layers} & \textbf{GFLOPS} & \textbf{Top-1} \\
\hline
ResNet50-3D \cite{hara2018can} & - & 50 & 80.32 & 0.613\\
\hline
ResNet101-3D \cite{hara2018can} & - & 101 & 110.98 & 0.652\\
\hline
ResNeXt101-3D \cite{hara2018can} & - & 101 & 148.91 & 0.651\\
\hline
Wide ResNet50-3D \cite{hara2018can} & - & 50 & 72.32 & 0.639\\
\hline
\hline
I3D \cite{carreira2017quo} & ImageNet & 48 & 55.79 & 0.664 \\
\hline
MF-Net \cite{chen2018multifiber} & ImageNet & 50 & 22.7 & 0.728 \\
\hline
R(2+1)D-ResNet50 \cite{tran2018closer} & Sports1M & 50 & 238.12 & 0.720  \\
\hline
\hline
ResNet101-3D (w/ \textit{ClassReg}) & - & 101 + 4 & 126.13 & 0.677\\
\hline
Wide ResNet50-3D (w/ \textit{ClassReg}) & - & 50 + 4 & 82.67 & 0.653\\
\hline
I3D (w/ \textit{ClassReg}) & - & 48 + 3 & 62.96 &  0.678\\
\hline
\end{tabular}}
\end{center}
\vspace{-7mm}
\end{table}

\subsection{Direct comparisons with Class Regularization}
\begin{figure}[ht]
\centering
\includegraphics[width=.9\linewidth]{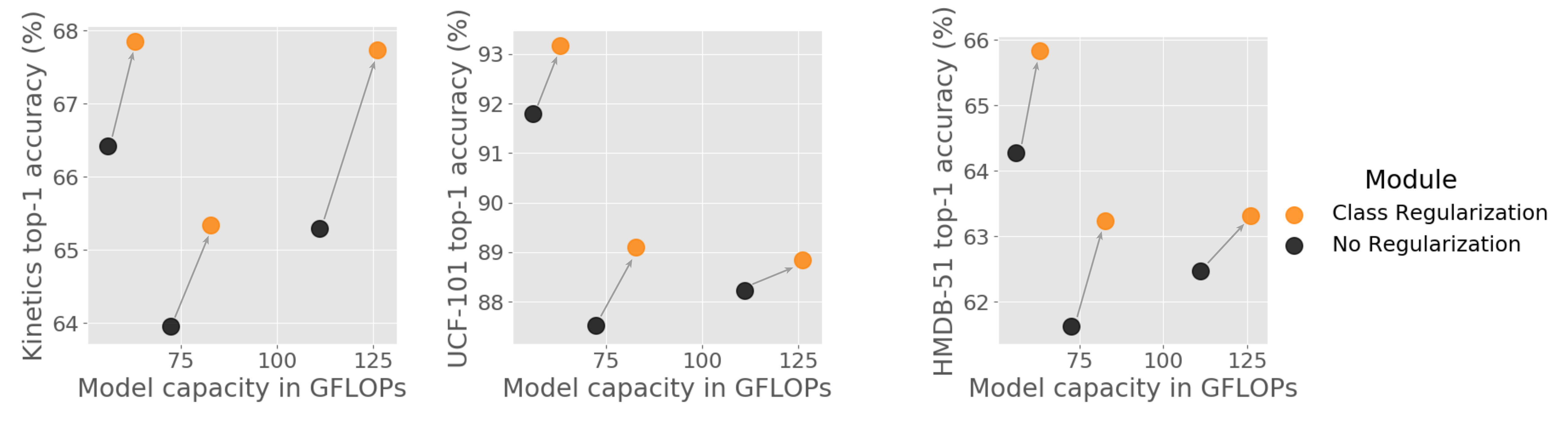}
\caption{Class Regularization accuracy/computation trade-off. Clip top-1 accuracy for the Kinetics, UCF-101 and HMDB-51 in comparison to the computational cost (in GFLOPs).}
\label{fig:accuracy_to_flops}
\vspace{-5mm}
\end{figure}

% Results on Kinetics
In Table~\ref{table:table2} we compare the \textit{Class Regularization} method in a \textit{per-architecture} fashion by keeping a base network and reporting accuracy rates in pairs. For each architecture and dataset, networks with \textit{Class Regularization} outperform those without. The largest gain was observed in the 101-layer 3D Resnet with a gain of +2.45\% on Kinetics, +0.61\% on UCF-101 and +0.81\% on HMDB-51. On Wide-Resnet50 we further obtained improvements of +1.37\%, +1.59\% and +1.62\% for each of the respected datasets. On I3D \textit{Class Regularization} provided an increase of +1.43\% for Kinetics, +1.37\% on UCF-101 and +1.56\% on HMDB-51. This is also based on the direct correlation between the complexity of the class features in the prediction layer given the architectural depth. Since the effective description of classes is done through large feature spaces, \textit{Class Regularization} can significantly benefit models that include highly complex and large class weight spaces. With this, the corresponding set of influential class features are being better distinguishable with minimal computational costs as seen in Figure~\ref{fig:accuracy_to_flops}

% Class Regularization with Transfer Learning
%When transferring weights, the re-training process does not change for the \textit{Class Regularization} networks as the additional training phase is only performed in order for the model to learn the feature correspondence.

%Evidence that advancements can also be achieved in transfer learning are seen by the rates attained in both UCF-101 and HMDB-51 in Table~\ref{table:table2}, with rates from non-\textit{Class Regularized} networks being re-calculated in order to include the same training setting. The largest gain from the original implementation in UCF-101 was on the Wide ResNet50 with +1.59\% followed by I3D with +1.26\% and ResNet101 +0.61\%. Similarly, in the HMDB-51 dataset, the model pair that exhibited the largest gap in performance was Wide ResNet50 with a +1.62\% improvement, I3D with +1.56\% and ResNet101 with +0.84\%. Overall, the minor deterioration of the accuracy gains could be contributed to the fact that kernels have been already trained in conjunction with class information from a different dataset.  

\begin{table}[htb]
\caption{Direct comparison with and without regularization block. Models that include \textit{Class Regularization} are in \textcolor{orange}{orange}. Reported accuracy rates (top-1 \%) achieved on Kinetics-400, UCF-101 and HMDB-51 datasets on the validation sets (split 1 for UCF-101 and HMDB-51), with all networks being re-trained with the same settings. All models use inputs of size $16 \times 112 \times 112$ for Kinetics and $16 \times 224 \times 224$ for UCF-101 and HMDB-51. Initially, all networks are trained for 170 epochs. During fine-tuning, we trained for 100 epochs.\label{table:table2}}
\vspace{-6mm}
\begin{center}
\resizebox{\columnwidth}{!}{%
\begin{tabular}{l|c|c|c|c}
\hline
\textbf{Model} & \textbf{added latency(msec.)} & \textbf{Kinetics} & \textbf{UCF101} & \textbf{HMDB51}\\
\hline
ResNet101  & - & 65.29 & 88.23 & 62.47\\
\hline
\color{orange} ResNet101  & + 98.786  & 67.74 & 88.84 & 63.31\\
\hline\hline
Wide ResNet50 & -  & 63.96 & 87.52 & 61.62\\
\hline
\color{orange} Wide ResNet50  & +102.995 & 65.33 & 89.11 & 63.24\\
\hline\hline
I3D & - & 66.42 & 91.80 & 64.27\\
\hline
\color{orange} I3D & +68.340  & 67.85 & 93.17 & 65.83\\
\hline\hline
\end{tabular}}
\end{center}
\vspace{-7mm}
\end{table}

\section{Conclusions}
\label{sec:sec5}

% General premise
%Convolutions are performed over multiple number features in each layer while not having a direct association between layer feature activations and task classes.

% Class Regularization
In this paper, we have introduced \textit{Class Regularization}, a method that focuses on class-specific features. \textit{Class Regularization} allows the network to strengthen or weaken layer activations based on how informative they are to specific class predictions. The method can be added to any layer or block of convolutions in pre-trained models. It is lightweight as the class weights from the prediction layer are shared throughout \textit{Class Regularization}. To avoid the vanishing gradient problem, and the possibility of negatively influencing activations, the weights are normalized between a range given an \textit{affection rate} ($\alpha$) value.

% Evaluation
We evaluate the proposed method on three benchmark datasets: Kinetics, UCF-101 and HMDB-51 and report results on three models: ResNet101, Wide ResNet50 and I3D with average increases in accuracy of +1.29\%, +1.5\% and 1.45\% respectively. In addition, the achieved improvements were done with minimal additional computational cost over the original architectures.

% Additional applications
We demonstrate how \textit{Class Regularization} can be used in order to improve explainability of 3D-CNNs through qualitative class feature visualizations across layers, and quantitative class predictions improvements for different layer depths.

\section{Acknowledgments}
This publication is supported by the Netherlands Organization for Scientific Research (NWO) with a TOP-C2 grant for ``Automatic recognition of bodily interactions'' (ARBITER).

\bibliographystyle{IEEEbib}
\bibliography{refs}

\end{document}